\documentclass[letterpaper, 10 pt, conference]{ieeeconf}  
\IEEEoverridecommandlockouts                              
\usepackage{amsmath}

\usepackage{amsthm}
\usepackage{amssymb}
\usepackage{enumerate}
\usepackage{hyperref}
\hypersetup{
    colorlinks=true,
    linkcolor=blue,
    filecolor=magenta,      
    urlcolor=cyan,
}
\usepackage{bm}
\usepackage[pdftex]{graphicx}
\usepackage{subcaption}
\usepackage[inline]{asymptote}
\usepackage[normalem]{ulem}
\usepackage{booktabs}
\usepackage{textcomp}
\usepackage{tabularx}
\usepackage{array}
\usepackage{cite}
\useunder{\uline}{\ul}{}

\usepackage{bm}

\newcommand{\R}{\mathbb{R}} 
\newcommand{\vect}[1]{\boldsymbol{\mathbf{#1}}} 

\DeclareMathOperator*{\argmin}{arg\,min}

\newcommand{\norm}[1]{\left\lVert#1\right\rVert}

\newcommand{\smalltitle}[1]{\vspace{0.2em}\noindent \textbf{{#1}}}
\newcommand{\eg}{\textit{e.g.}}
\newcommand{\ie}{\textit{i.e.}}
\newcommand{\TR}{\vect{\xi}^{\text{R}}}
\newcommand{\TA}{\vect{\xi}^{\text{A}}}
\newcommand{\q}{\vect{q}}

\newcommand{\PreserveBackslash}[1]{\let\temp=\\#1\let\\=\temp}
\newcolumntype{C}[1]{>{\PreserveBackslash\centering}p{#1}}
\newtheorem{theorem}{Theorem}[section]
\newtheorem{lemma}[theorem]{Lemma}

\begin{document}
\title{ \LARGE \bf
Collaborative Interaction Models for Optimized Human-Robot Teamwork}


\author{
Adam Fishman$^{1,2}$
Chris Paxton$^{1}$,
Wei Yang$^{1}$, 
Dieter Fox$^{1,2}$
Byron Boots$^{1,2}$,
and Nathan Ratliff$^{1}$,
\thanks{$^{1}$ NVIDIA {\small \tt \{cpaxton, wyang, nratliff\}@nvidia.com}}
\thanks{$^{2}$ University of Washington {\tt \small \{afishman, fox, }}
\thanks{~~~~~~~~~~~~~~~~~~~~~~~~~~~~~~~~~~~{\small \tt bboots\}@cs.washington.edu}}
}
\maketitle
\begin{abstract}
Effective human-robot collaboration requires informed anticipation.  The robot must anticipate the human's actions, but also react quickly and intuitively when its predictions are wrong. The robot must plan its actions to account for the human's \emph{own} plan, with the knowledge that the human's behavior will change based on what the robot actually does. This cyclical game of predicting a human's future actions and generating a corresponding motion plan is extremely difficult to model using standard techniques. In this work, we describe a novel Model Predictive Control (MPC)-based framework for finding optimal trajectories in a collaborative, multi-agent setting, in which we simultaneously plan for the robot while predicting the actions of its external collaborators. We use human-robot handovers to demonstrate that with a strong model of the collaborator, our framework produces fluid, reactive human-robot interactions in novel, cluttered environments. Our method efficiently generates coordinated trajectories, and achieves a high success rate in handover, even in the presence of significant sensor noise. See our full video at \url{ https://youtu.be/bSY8K-jkRtA } for a summary of our method, as well as videos of our experiments and real-robot trials.
\end{abstract}

\section{Introduction}

Human behavior is determined by a mixture of intent, world prediction, anticipation, physical limitations, and more. When planning in the presence of people, robotic decision processes often encapsulate these diverse desiderata under the lid of a black box dynamics function.
When the robot and human's goals are independent~\cite{ZiebartPedestrians2009,MainpriceHumanRobotReaching2016,bai2015intention}, this model has been very successful.

\begin{figure}[bt]
\centering
\includegraphics[width=\columnwidth]{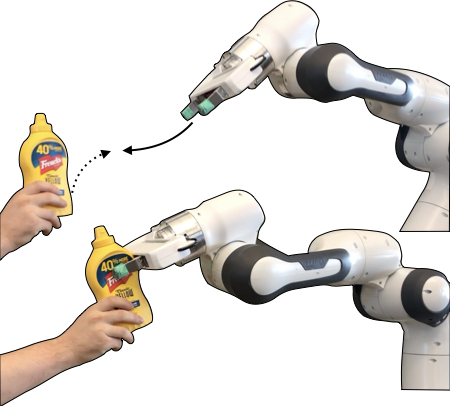}
\captionsetup{size=footnotesize}
\vspace{-.1in}
\caption{Coordinating a fluid human-robot handover requires an estimate of the human's plan, so that the robot can be in position to make the handover at the correct time. Our algorithm can achieve smooth and natural human-robot collaborative motions in a variety of scenarios, even in the presence of obstacles and sensor uncertainty.
}
\vspace{-0.2in}
\label{fig:cover}
\end{figure}

However, cooperating to achieve shared goals is more difficult. 
Take the human-robot hand-over task shown in Fig.~\ref{fig:cover} as an example, where a robot must receive some object from a human collaborator. 
Humans will act based on what they imagine the robot will do~\cite{dragan2013legibility}, and,
conversely, the robot should choose actions based on 
its best estimate of the human's intention. 
Predicting human intention while planning is not new, this has been explored in anticipatory planning~\cite{koppula2016anticipatory}, and prior work has modeled human kinematics and dynamics in order to achieve collaborative manipulation tasks
~\cite{toussaint2017multi,peternel2017towards,vogt2018one}.

However, humans often do not act according to plan. Any robot planner that tries to predict their intentions must be highly reactive. 
We propose a Model Predictive Control (MPC) approach, which models both human and robot as separate, fully-actuated actors in a combined trajectory optimization problem. Our MPC approach allows the robot to determine the most effective form of collaboration while still being able to react to changing circumstances and noisy sensor data.
We specifically apply our approach to the problem of human-robot handover~\cite{Strabala2013TowardSH,Huang2015AdaptiveCS,Maeda2017ProbabilisticMP,Medina2016AHC}.
The core problem is that the human and robot must coordinate on where and how the handover is to take place~\cite{Strabala2013TowardSH}.
In effect, we must balance between the two cost functions for human and robot, avoiding obstacles while finding the most logical location for both to reach.

This combined human-robot system is both partially observable and under-actuated since the robot has no real control over the human and cannot directly observe factors influencing their decisions. Therefore, at a minimum, our planner must be \emph{reactive} \cite{MurphyPOMDPSolvers2000} to unforeseen human behavior.
 We follow a real-time Model-Predictive Control (MPC) paradigm and re-optimize with each new observation. 
Computation speed is also crucial. We employ a modern motion optimization strategy, which leverages fast Gauss-Newton solvers \cite{Mukadam2018ContinuoustimeGP, RatliffRieMO2015ICRA,17-toussaint-Newton}, and assume relevant aspects of the human and robot are fully actuated.

Additionally, to ensure spatial consistency of the resulting reactive behavior through re-optimization, we introduce a novel class of explicit sparse reward terms, \ie, negative costs, around the target. 
Within a certain radius, the robot is explicitly rewarded for approaching the target, thus extending the target's influence beyond a terminal potential to each intermediate time step. 
The system is therefore able to compromise between goal accumulation and trajectory smoothness.


We evaluate our technique in both simulated ablation studies as well as real-world handovers between a human participant and a Franka robot using a real-time perception system. We show that, especially in the presence of obstacles, our technique enables the robot to anticipate the human's actions leading to well-coordinated, quick, and smooth handover behavior while timing the handover better than the alternatives, both quantitatively and qualitatively.

\section{Related Work}
Modeling human behavior is crucial for successful human-robot collaborative manipulation and has been explored in a variety of contexts~\cite{bai2015intention,Maeda2017ProbabilisticMP,zhou2018early,toussaint2017multi}.
In addition, many recent methods for human-robot handover use perception and some manner of human modeling to achieve reactivity~\cite{peternel2017towards,Maeda2017ProbabilisticMP,kupcsik2018learning}.
However, these models are usually uni-directional, with information flowing from prediction to planner but not vice versa.
For example, Ziebart et al.\cite{ZiebartPedestrians2009} used predicted goal-oriented pedestrian behavior to augment navigation planners to minimize interaction. Similarly, Mainprice et al.~\cite{MainpriceHumanRobotReaching2016} modeled human reaching behaviors to reduce interaction or collision events while working along-side humans.
Maeda et al.~\cite{Maeda2017ProbabilisticMP} use probabilistic motion primitives to model both humans and robots in a variety of collaborative tasks, including handover. Zhou et al.~\cite{zhou2018early} used a recurrent neural net to model human activity for collaboration in the operating room.

In reality, however, the human will respond to qualities of the robot's motion, \eg, speed and shape, trying to estimate and adapt to its motion. Humans and robots can collaborate more effectively if the robot's motions appear legible to humans~\cite{dragan2013legibility} and allow the humans to understand the robot's goals~\cite{grigore2013joint}.
One way to achieve legibility is to use human demonstration data to teach the robot~\cite{Maeda2017ProbabilisticMP}. 
Another approach focuses on jointly modeling the human-robot system as some sort of hybrid planning problem,~\cite{bai2015intention,toussaint2017multi,stouraitis2018dyadic}, and try to structure the problem to ensure effective collaboration.


Human-robot handovers are a particularly well-studied area for human-robot collaboration, with applications both to industry~\cite{peternel2017towards,unhelkar2014comparative} and to in-home assistance~\cite{Huang2015AdaptiveCS}. 
Much prior work analyzes the formulation of the human handover and how to structure the action naturally \cite{Strabala2013TowardSH,Huang2015AdaptiveCS}. This can be particularly well represented as a hybrid planning problem~\cite{toussaint2017multi,stouraitis2018dyadic}.
Toussaint et al.~\cite{toussaint2017multi} proposed a method for offline planning based on Task and Motion Planning. This allows for longer-horizon planning across grasps as compared to our method, but is inherently less reactive.
Other work used a dyadic model for collaboration between a human and a robot~\cite{stouraitis2018dyadic}.

Our method applies more specifically to the approach phase of the handover. 
Related ideas include exploiting a database of human demonstrations to produce natural and fluid plans~\cite{Yamane2013SynthesizingOR}. Likewise, Maeda et al.~\cite{Maeda2017ProbabilisticMP} use imitation learning to mimic human behavior and Medina et al.~\cite{Medina2016AHC} use a human-inspired dynamics controller to model the entire action: approach, grasp, retract. These methods are well-suited for controlled interaction settings, but generalizing them to handle the diversity of speed and environmental variations encountered in the real world is challenging. 
Peternel et al.~\cite{peternel2017towards} also model the human during collaborative manipulation, but their goal is to minimize risk of injury, whereas our goal is to achieve fluid collaboration in the presence of obstacles.

Our work relies on motion optimization approaches that are both fast and expressive~\cite{17-toussaint-Newton,RatliffRieMO2015ICRA,Mukadam2018ContinuoustimeGP}. Motion planning as an optimization problem was first presented in \cite{RatliffCHOMP2009}, and accelerated in a quick progression of work \cite{KalakrishnanSTOMP2011,ParkITOMP2012,SchulmanTrajopt2013}. These early optimizers addressed primarily the subproblem of smooth collision avoidance. The work of \cite{ToussaintTrajOptICML2009,Toussaint2014NewtonProblems,RatliffRieMO2015ICRA}, extended the paradigm showing that generic second-order Gauss-Newton optimizers out-of-the-box could solve a more general class of constrained motion optimization problem.
Soon thereafter, Mukadam et al.~\cite{Mukadam2018ContinuoustimeGP} demonstrated that standard factor graph tools could drastically simplify the modeling. We build on these ideas here, using a factor graph to model the problem and fast modern optimizers to solve the continuous optimization loop in real time.

While our setting is fundamentally partially observable, we do not address the Partially Observable Markov Decision Process (POMDP) problem directly, other than to use standard, reactive maximum a posteriori (MAP) approximation techniques~\cite{MurphyPOMDPSolvers2000} to motivate the importance of continuous re-optimization. 
Using maximum-likelihood observations and active replanning has proven useful before, even for very complex multi-stage tasks~\cite{garrett2019online}.
Other approaches use Monte Carlo sampling to explore possible outcomes for various actions~\cite{silver2010monte,somani2013despot,bai2015intention}.
Some POMDP work has even actively modeled uncertainty over human intention~\cite{bandyopadhyay2013intention,bai2015intention}, particularly in the context of autonomous vehicles~\cite{bai2015intention}.

\section{Theoretical framework}\label{sec:theory}


In games, agents try to optimize individual objectives \cite{AlgorithmicGameTheory2007}, but 
collaborative tasks require cooperation. When collaborating, agents collectively
optimize a single system objective. In this section, we formalize the 
collaborative system. We derive
predictive models for each external, \ie, uncontrolled, agent and an optimal control objective
for the controlled agent, \ie the robot. We then show that if the models of the external agents' behavior are
sufficiently predictive, the controlled agent can achieve a stable \textit{collaborative equilibrium} by choosing actions
according to its objective. 
 
We consider a system constituting $N+1$ total agents. Let the $0^{\text{th}}$ agent denote the controlled agent and agents $1, \ldots, N$ be external, uncontrolled, but collaborating agents.

\subsection{Collaborative interaction model}


Denote the $i^{\mathrm{th}}$ agent's trajectory by 
$\vect{\xi}_i = (\q_i^{0},\q_i^{1},\cdots,\q_{i}^{T+1})$. 
Let $\vect{\varsigma}_i^{t} = (\q_{i}^{t-1}, \q_{i}^{t}, \q_{i}^{t+1})$
denote the trajectory's $t^\mathrm{th}$ second-order clique,\footnote{Here we use superscripts 
just for notational convenience of time indexing, not to be confused
with the component indexing of tensor notation.} a triple of consecutive positions 
used to represent position and the corresponding finite-difference approximations of velocity,
and acceleration at each time step \cite{Toussaint2014NewtonProblems}.

We define the joint \textit{collaborative system trajectory} as $\vect{\xi}=(\vect{\xi}_0,\vect{\xi}_1,\cdots,\vect{\xi}_N)$ and denote
its constituent 2nd-order cliques  by
$\vect{\varsigma}^t = (\vect{\varsigma}_0^t, \vect{\varsigma}_1^t, \ldots, \vect{\varsigma}_N^t)$.

Denoting the space of all collaborative system trajectories by $\Xi$, we define the system's \textit{collaborative interaction model} (or simply its \textit{collaboration model})
$\mathcal{M}=\{C,G,H\}$ as
\begin{eqnarray} \label{eq:CollaborativeInteractionModel}
    \min_{\vect{\xi}} C(\vect{\xi}) \quad \mathrm{s.t.} \quad G(\vect{\xi}) \leq 0, ~H(\vect{\xi}) = 0
\end{eqnarray}
where $C:\Xi\rightarrow\R$ is the collaborative cost, $G:\Xi\rightarrow\R^k$
are $k$ inequality constraint functions, and $H:\Xi\rightarrow\R^l$ are $l$
equality constraint functions.
For compactness, we use $\Xi_\mathcal{M}\subset\Xi$ to denote the feasible set of trajectories
that satisfy the constraints $G$ and $H$.
We can then write the collaborative interaction model as 
$\vect{\xi}^* = \argmin_{\vect{\xi}\in \Xi_\mathcal{M}} C(\vect{\xi})$. 
The model for a given collaborative system can change incrementally over time as the environment, the agents' goals, or the agents themselves change. 
We assume that the optimizer is able to track solutions over time within a continuous optimization loop, such as Model Predictive Control (MPC). Note that both the costs / constraints and the set of available trajectories $\Xi_M$ usually change from cycle to cycle updated with the latest estimates of the world and agent states.

In our experiments, we export a $k^\mathrm{th}$-order Markov structure
in the system \cite{Toussaint2014NewtonProblems} enabling us to 
write the collaborative interaction model of Equation~\ref{eq:CollaborativeInteractionModel} in clique notation as
\begin{align}
    &\min_\xi \sum_{t=1}^T c_t(\vect{\varsigma}^t) 
    \ \ \mathrm{s.t.}\ \  g_t(\vect{\varsigma}^t) \leq 0, ~h_t(\vect{\varsigma}^t) = 0
    \ \ \forall t. 
\end{align}
Often, more complex task spaces are defined on these cliques by transforming them through differentiable task maps where objective terms may reside. It is common to represent the task spaces, \ie, the co-domains of the task maps, with maximal coordinates, which are an explicit representation of the task space constrained to match the output of the task map. Following this paradigm, we define our optimization costs and constraints in maximal coordinates on the relevant task space. We also use soft constraints implemented as unconstrained penalties in our experiments as in \cite{Dellaert2017FactorPerception}.
In this section, though, we use the more compact notation 
given above for succinctness and generality.

The key intuition behind our model is that although we cannot explicitly
control the $N$ external collaborating agents, we assume we can sufficiently
predict their behavior and treat prediction errors as system disturbances. 
We make this assumption concrete below and explore it
experimentally in Sec.~\ref{sec:ModelingHuman}.

When the 
collaborative interaction model has a unique global minimum, that minimizer acts as an equilibrium point and becomes predictable by the agents in a way we can exploit in our model (explored below in Section~\ref{sec:MutualPredictability}). We, therefore, call the global minimum the system's \textit{collaborative equilibrium} and say the system is well-defined if it has a unique global minimum.
In this work, we assume both that the global minimum is well defined and that an optimizer will be able to track the global minimum over time. In practice, these assumptions amount to the agents mutually knowing the higher-level collaboration plan either in advance or by sufficiently communicating it to each other unambiguously on the fly. For complex tasks, there may be many local minima or even regions of equally good global minima, representing different equilibria. In these cases, the system would require additional estimation machinery to maintain predictive distributions across external agent behavior, which we do not address here.

We start by defining explicitly the agents' individual
predictive \textit{collaborative behavior models} implicit in 
the above collaborative interaction model. 
Let $\Xi_\mathcal{M}[\vect{\xi}_i]$ denote the feasible set of system 
trajectories where the $i^{\mathrm{th}}$ agent's trajectory is fixed at $\vect{\xi}_i$.
We define the $i^\mathrm{th}$ agent's \textit{predictive collaborative cost} to be
\begin{align}
    c_i(\vect{\xi}_i) = \min_{\vect{\xi}^{\backslash i}\in\Xi_\mathcal{M}[\vect{\xi}_i]} C(\vect{\xi}_i, \vect{\xi}^{\backslash i}),
\end{align}
where with a slight abuse of notation, we use $C(\vect{\xi}_i, \vect{\xi}^{\backslash i})$ to denote the collaborative cost evaluated at the joint system trajectory defined by agent $i$'s trajectory $\vect{\xi}_i$ and the remaining system trajectories $\vect{\xi}^{\backslash i}$ of all other agents $j\neq i$.
This cost encodes the agent's action criteria under an 
assumption that all other agents are predicted as having optimal collaborative
responses under the system's collaboration model. Note that these predictive models are assumed to know agent $i$'s intent (the trajectory $\vect{\xi}_i$). While this assumption is generally wrong, as the robot cannot truly know an external agent's intent, we will see below that it is valid at the system's collaborative equilibrium where equilibrium behavior becomes mutually predictable (see Section~\ref{sec:MutualPredictability}).

Each agent then has its own individual \textit{collaborative behavior model} of the form
\begin{align}
    \vect{\xi}_i^* = \argmin_{\vect{\xi}_i\in\Xi_\mathrm{M}^i} c_i(\vect{\xi}_i),
\end{align}
where $\Xi_\mathrm{M}^i$ is the set of all trajectories $\vect{\xi}_i$ for the
$i^\mathrm{th}$ agent for which $\Xi_\mathrm{M}[\vect{\xi}_i]$ is nonempty,
\ie, $\Xi_\mathrm{M}^i = \{\vect{\xi}_i\ |\ \Xi_\mathrm{M}[\vect{\xi}_i] \neq \emptyset\}$

\subsection{Stability of the predictive controller}

We adopt definitions of stability from control theory and say that an assignment of behavior generation algorithms to the agents are collectively, or asymptotically, stable around the equilibrium if the joint system evolves stably, or stably asymptotically, around the system trajectory. Under this notion of stability, we can make following statement. 
\begin{lemma}
Suppose we have a collaborative system and a corresponding collaboration model $\mathcal{M}$. If we can say that a MPC algorithm over $\mathcal{M}$
rejects $\epsilon$-disturbances and that each agent's collaborative behavior model is 
$\epsilon$-predictive of the agent's next action, including the controlled 
agent's execution under the environment's stochasticity,
then controlling the controlled agent with the MPC algorithm will create system behavior that is 
stable around the collaborative equilibrium of $\mathcal{M}$.
\end{lemma}
In other words, if our collaboration model is sufficiently predictive for the external agents and we control our controllable agent using the collaborative behavior model derived from it, the combined system behavior is stable around the collaborative equilibrium.

Note that in this stability statement, the metric $\epsilon$-predictive is undefined. This is because the statement will hold as long as the definition of $\epsilon$-predictive is consistent with the definition of $\epsilon$-disturbances, \ie, the range of system deviations that can be handled by MPC. While we cannot explicitly control the external agents, if we can predict their behavior sufficiently well, then we may treat deviations as system disturbances. With this, we do not need to assume that the external agents generate behavior with the same collaborative system model, as the collaborative behavior models induced by the system are sufficiently predictive.

\subsection{Mutual predictability of the collaborative equilibrium}
\label{sec:MutualPredictability}

Each collaborative behavior model implicitly uses a \textit{conditional} model to predict the behavior of
external agents. Specifically, under agent $i$'s collaborative behavior 
model, the cost $c_i(\vect{\xi}_i)$ optimizes over each external agent $j\neq i$ \textit{given} agent $i$'s
trajectory $\vect{\xi}_i$, thus modeling the response the other agents would have if they
were given knowledge of $\vect{\xi}_i$. The model assumes that all responding agents know the agent $i$'s intent, which is in general
not true. However, equilibrium behavior has a mutual predictability property which enables all agents behavior to be predictable, thereby validating the conditional model specifically at the collaborative
equilibrium. 

Equilibrium predictions of other agent's behavior are both reflexive and transitive, creating a stationarity property of the predictions. For instance, let $\vect{\xi}_j^\mathcal{M}(\vect{\xi}_i)$ be the implicit prediction made by 
agent $i$ of how agent $j$ will respond to agent $i$'s intended trajectory $\vect{\xi}_i$. Let $\vect{\xi}^* = \{\vect{\xi}_i^*\}_{i=1}^N$ denote the collaborative 
equilibrium of system $\mathcal{M}$. Then for all $i,j$ we have $\vect{\xi}_j^\mathcal{M}(\vect{\xi}_i^*) = \vect{\xi}_j^*$. Therefore, 
$\vect{\xi}_i^\mathcal{M}(\vect{\xi}_j^\mathcal{M}(\vect{\xi}_i^*)) = \vect{\xi}_i^*$ (reflexive) and $\vect{\xi}_k^\mathcal{M}(\vect{\xi}_j^\mathcal{M}(\vect{\xi}_i^*)) = \vect{\xi}_k^* = \vect{\xi}_k^\mathcal{M}(\vect{\xi}_i^*)$ (transitive).

In other words, each agent
predicts an equilibrium response under equilibrium behavior. Even though the agent uses a conditional predictive
model which assumes external agents know the agent's intent, 
specifically at the collaborative equilibrium,
the agent's intended behavior becomes predictable as part of the equilibrium behavior validating the use of the conditional model.

\section{Human-Robot Handover using Finite-Horizon Optimization}
In this section, 
we formulate the handover task as an application of our general framework where the collaborative model optimizes for the human and robot successfully reaching each other to perform the handover. We detail the objective terms used in our models (Sections~\ref{sec:ModelingRobot}, \ref{sec:ModelingHuman}, \ref{sec:ModelingInteraction}) and discuss implementing spatially consistent behavior using finite-time-horizon MPC (see Section~\ref{sec:TimeIndependence}). 

In this section, we consider the robot to be the controlled agent (agent $0$) and the human to be the uncontrolled external agent (agent $1$), and denote their trajectories as $\TR = \{\vect{q}^\text{R}_i\}_{i=0}^T$ (robot) and $\TA = \{\vect{q}^\text{A}_i\}_{i=0}^T$ (external agent, \ie human), respectively. We focus here on defining the unconstrained objective, making the common assumption that many constraints can be naturally modeled as soft constraints using fixed penalties (see, for instance, \cite{Dellaert2017FactorPerception}). This is a reasonable approximation, especially since stochasticity makes the optimization inherently approximate.

The collaboration objective can be decomposed into three terms, a robot specific term, an external agent (human) specific term, and an interaction term. 
\begin{align}
    C(\xi) = \lambda_\text{R} c^\text{R}(\TR) 
    + \lambda_\text{A} c^\text{A}(\TA) 
    + \lambda_\text{I} c^\text{I}(\TR, \TA)
\end{align}
We detail these three terms in the following sections.

\subsection{Modeling the robot} \label{sec:ModelingRobot}


First, we define the cost modeling the robot trajectory, 
\begin{eqnarray}\label{eq:cost-robot}
    c^\text{R}(\vect{\xi}^\text{R}) = \sum_{i=0}^T c^\text{R}(\vect{q}_i^\text{R}, \dot{\vect{q}}_i^\text{R}, \ddot{\vect{q}}_i^\text{R}),
\end{eqnarray}
where $T$ is the total number of time steps and $\vect{q}^\text{R}$, $\dot{\vect{q}}^\text{R}$, $\ddot{\vect{q}}^\text{R}$ are the position, velocity, and acceleration of the joints of the robot in configuration space. In what follows, we will also use $\vect{x}_i^\text{R} = \phi(\vect{q}_i^\text{R}) = [\vect{R}_i^\text{R}, \vect{t}_i^\text{R}]$ to represent the 6-DOF pose of the end effector in the world frame, after applying the forward kinematics function $\phi(\cdot)$.

Equation \ref{eq:cost-robot} can be split into the sum of individual cost functions, which we define in the following sections.

\smalltitle{Obstacle avoidance and joint constraints. } 
To prevent hitting the joint limits and to avoid obstacles, we include three cost functions 
$c_{\text{joint}}(\vect{q}_i^\text{R})$, 
$c_{\text{joint}}(\dot{\vect{q}}_i^\text{R})$,
and $c_{\text{obs}}(\vect{q}_i^\text{R})$. 

Let $J$ denote the indices of the joints, $\theta_j^\text{R}$ denote the angle of $j$th joint, and $(\theta_{\text{j, min}}^\text{R}, \theta_{\text{j, max}}^\text{R})$ denote the corresponding joint limitation, we employ a hinge-loss-based cost~\cite{Mukadam2016GaussianPlanning} for the joint limit:
\begin{eqnarray}
    c_{\text{joint}}(\vect{q}_i^\text{R}) &= \sum_{j \in J} \left\lVert c(\theta_j^\text{R})\right\rVert^2,
\end{eqnarray}
where $c(\theta_j^\text{R})$ is defined as
\begin{eqnarray} \label{eq:joint-cost}
    c(\theta_j^\text{R}) &= \begin{cases}
-\theta_j^\text{R} + \theta_{\text{j, min}}^\text{R} - \epsilon_j, & \text{if } \theta_j^\text{R} < \theta_{\text{j, min}}^\text{R} + \epsilon_j \\
\theta_j^\text{R} - \theta_{\text{j, max}}^\text{R} + \epsilon_j, & \text{if } \theta_j^\text{R} > \theta_{\text{j, max}}^\text{R} - \epsilon_j \\
0, & \text{otherwise}
\end{cases}
\end{eqnarray}
Here $\epsilon_j$ is the joint limit error tolerance for joint $j$.

We also impose a cost 
\begin{eqnarray}
    c_{\text{joint}}(\dot{\vect{q}}_i^\text{R}) &= \sum_{j \in J} \left\lVert c(\dot{\theta}_j^\text{R})\right\rVert^2,
\end{eqnarray}
where $c(\dot{\theta}_i)$ is formulated similarly to Equation \ref{eq:joint-cost} using $\dot{\theta}$ in place of $\theta$.

These cost functions assume that the environment is static, \ie, the camera and the obstacles are unchanging. Within the context of MPC, however, we are able to update these cost functions as the environment changes and we redefine our optimization problem. We compute a signed distance field representing a discretization of the environment. Then, as in \cite{RatliffCHOMP2009}, we use a sphere-based ``skeleton" that covers the robot's entire volume and surface area. The spheres allow for a sparse and efficient representation of the robot's volume. Our total obstacle cost is then the sum of the cost at each sphere:
\begin{eqnarray}
c_{\text{obs}}(\vect{q}^\text{R}_i) = \sum_{s \in \text{spheres}} \left\lVert c(s)\right\rVert^2, 
\end{eqnarray}
where
\begin{eqnarray}
c(s) &= \begin{cases}
-d_s + s_\text{radius}, & \text{if } d_s \leq s_\text{radius} \\
0, & \text{if } d_s > s_\text{radius}
\end{cases}.
\end{eqnarray}
Here $d_s$ is the value of the signed distance function at the sphere $s$'s center.

\smalltitle{Velocity and acceleration constraints. } 
We include independent constraints for configuration-space velocity and acceleration constraints, $c(\dot{\vect{q}}^\text{R})$ and $c(\ddot{\vect{q}}^\text{R})$, that each constrain these values to be zero. The velocity penalty ensures that the robot slows after after reaching its goal, while the acceleration penalty ensures the robot moves fluidly without overshooting its target. 

\begin{figure}[t] 
\centering
\vspace{-25pt}
\includegraphics[width=0.9\columnwidth]{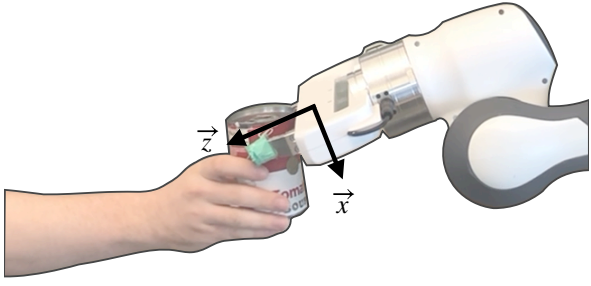}
\vspace{-25pt}
\captionsetup{size=footnotesize}
\caption{The ideal orientation for the end effector. The $z$-axis points toward the human's wrist, which we are able to track with the Microsoft Azure Kinect, and the $x$-axis is as vertical as possible.}
\label{fig:orientation}
\vspace{-10pt}
\end{figure}
\smalltitle{End effector constraints. } 
We also constrain the robot's end effector orientation to match an optimal value and add this to our overall cost function. In the 2D case, this optimal value is straightforward: the robot should always be oriented towards the human. The optimal 3D orientation is more complex.

Let $G$ denote the coordinate frame of the gripper where the z-axis $\vect{z}_G$ points directly out from the gripper and the x-axis $\vect{x}_G$ points down perpendicular to the gripper. When the gripper is perfectly flat, $\vect{x}_G$ points straight down to the ground. We wish to align $\vect{z}_G$ with $\vect{v}$, the ray from the end effector to the human's hand. We also want the end effector to be approximately flat. We show this ideal configuration in Figure \ref{fig:orientation}. Assuming the world frame has $\vect{z}_W$ up, we want to find $\vect{x}_G$ that when expressed in the world coordinates, has the lowest $\vect{z}$ coordinate, \ie, in the world frame, 
\begin{eqnarray}
0 = \vect{z}_G \cdot \vect{x}_G = \frac{\vect{v}}{\left\lVert \vect{v} \right\rVert} \cdot [x_{\vect{x}_G}, y_{\vect{x}_G}, z_{\vect{x}_G}],
\end{eqnarray}
where $[x_{\vect{x}_G}, y_{\vect{x}_G}, z_{\vect{x}_G}]$ correspond to the world frame coordinates of $[1, 0, 0]_G$ in the gripper frame. We also know
$x_{\vect{x}_G} = 1 - \sqrt{y_{\vect{x}_G}^2, z_{\vect{x}_G}^2}.$

If the gripper does not point straight up, we can first solve for $z_{\vect{x}_G}$, then take the derivative with respect to $y_{\vect{x}_G}$ and set it to zero in order to find the $\vect{x}_G$ that points most-down. Then, $\vect{y}_G = \vect{z}_G \times \vect{x}_G$ and we can use these three axes to construct our desired rotation matrix 
$\hat{\mathbf{R}}$.

With $\hat{\vect{R}}$, we use the same cost function from \cite{Dong17arxiv} to constrain the robot to face this direction.
$$
c(\vect{R}_i^\text{R}) = \log(\hat{\vect{R}}^{-1}\vect{R}_i^\text{R})^\vee
$$
where $\log(\cdot)$ is the logarithmic map and $\vee$ is the operator that takes a skew-symmetric matrix to a vector.

As a simplifying assumption to improve planning efficiency, we only compute $\hat{\vect{R}}$ once at each planning cycle using current observation of both the robot and agent. Since we run many iterations closed-loop, the robot will continue to face towards the human's position.

\subsection{Modeling the human} \label{sec:ModelingHuman}

We use a reduced, but similar, set of cost functions for the external agent, \ie, the human, $c^{\text{A}}(\TA) = \sum_{i=0}^T c^{\text{A}}(\vect{q}_i^\text{A}, \dot{\vect{q}}_i^\text{A}, \ddot{\vect{q}}_i^\text{A})$. 
We model the human hand as a floating sphere and the parameters $\lambda_{\text{A}}$ are determined through a set of 29 recorded reach-to-point tasks. In the task of handover where the robot and human are similar heights, we propose that a floating sphere is a sufficient model for the human. For other collaborative tasks, such as handover at significantly different heights, the human's morphology and the kinematic feasibility of the task would be important.

\begin{figure}[t]
\centering
\begin{subfigure}[t]{0.49\columnwidth}
    \centering
    \raisebox{2.8ex}{
        \includegraphics[width=.9\columnwidth]{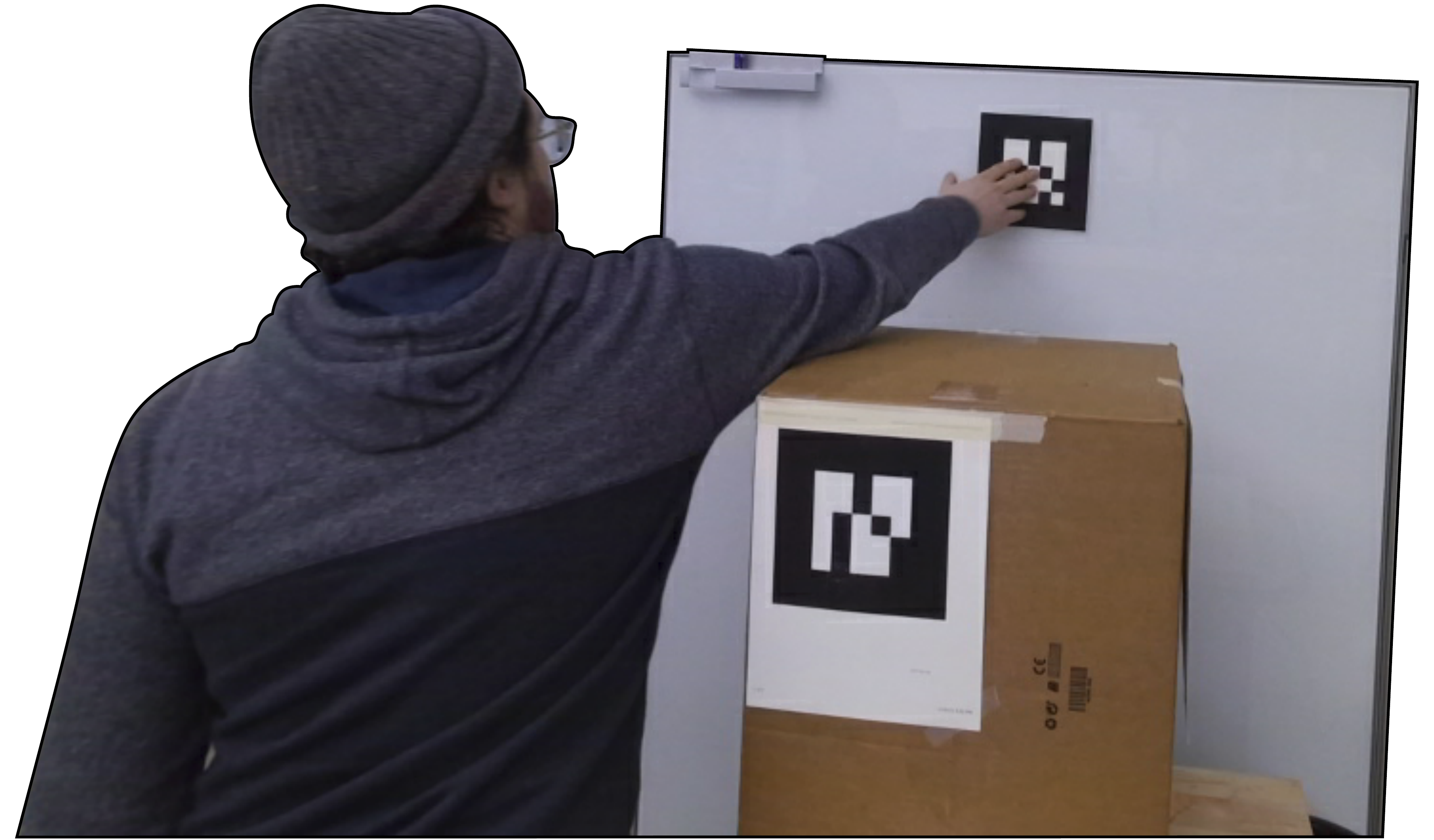}
    }
    \caption{}
\end{subfigure}
\begin{subfigure}[t]{0.49\columnwidth}
    \centering
    \includegraphics[width=.9\columnwidth]{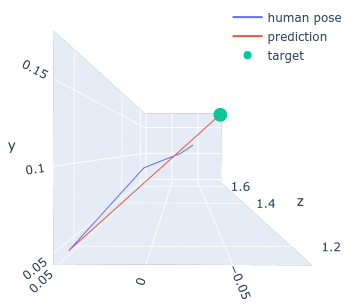}
    \caption{}
\end{subfigure}
\captionsetup{size=footnotesize}
\caption{(a) A reach-to-point task around an obstacle. We recorded 29 trials of a reach-to-point task, with varying target points, camera poses, and starting positions. (b) The predicted and measured human's trajectory for one trial. Here, the person chose to take a wider path around the obstacle than necessary.}
\vspace{-10pt}
\label{fig:human-prediction}
\end{figure}

Our model for optimal human reaching is a parametrized quadratic cost function that is nearly symmetrical to the robot. 
We model the human as a sphere representing their hand location, so we omit the constraint of joint limits. 
We also assume that a cooperative human will rotate their hand to meet the robot comfortably, and so we omitted the rotational constraint as well. 
To evaluate our model, we recorded a set of 29 reach-to-point tasks around an obstacle using the Microsoft Azure Kinect DK and its included body tracking SDK. Between trials, we randomly changed the target position, the human's starting pose, and the camera height. See Figure \ref{fig:human-prediction} for an example of our setup.

%

For each trial, we calculated our prediction error with 
\begin{eqnarray}
\text{Loss} = \sum_{t = 0}^{T}\frac{1}{T - (t + 1)} \sum_{i=t + 1}^{T}\norm{\vect{t}_{\text{predicted}} - \vect{t}_{\text{measured}}}^2
\end{eqnarray}
where $T$ is the total number of time steps it takes the human to reach the target and $x$ is the position of the hand. This loss represents the average distance between the corresponding real and predicted hand poses, which we then average over all MPC steps. We used $26$ of our trials to tune our human model and performed a grid search over $6,561$ parameter configurations to minimize error. We then evaluated the parameters on the remaining $3$ datasets. Our average loss on the training set $7.54$cm and our average loss on the evaluation set is $9.63$cm and a standard deviation of $2.41$cm.



\subsection{Modeling the robot-agent collaboration} \label{sec:ModelingInteraction}

At the end of the trajectory, the robot and the uncontrolled agent should meet. To enforce this, we encourage their end effector positions to be as close to each other as possible, 
\begin{eqnarray}
c^I(\vect{\xi}^\text{R}, \vect{\xi}^\text{A}) = 
\left\lVert \vect{t}^\text{R}_T - \vect{t}^\text{A}_T \right\rVert^2
\label{eq:robot-agent-collaboration-cost}
\end{eqnarray}
where $\vect{t}^{\text{R}}_T$ denotes the position of the robot's end-effector at the final time step and $\vect{t}^{\text{A}}_T$ denotes the position of the human's hand at that final time step (see the notation around forward kinematics in Section \ref{sec:ModelingRobot}).

The interaction term defines interaction only at the end of the trajectory (the behavior is finished once the interaction occurs). In general, it is unclear when (time-wise) this interaction should occur, so choosing a single $T$ is challenging, even more-so when re-optimizing the system and rejecting system perturbations within an MPC loop. The next section designs a sparse reward motivated by reinforcement learning settings to eliminate this problem, enabling spatially consistent behavior using a time-parameterized trajectory model.

\subsection{Time Independence through Sparse Rewards} \label{sec:TimeIndependence}
When two agents collaborate without explicit time synchronization, their interaction and behavior is often a function of combined state and not tied to a specific clock. For example, when handing over an object, both participants time their behaviors based on the observed state of the other, continually readjusting and aiming primarily to just meet in the middle.
The behavior is state-dependent and not explicitly timed.

We account for deviations from the planner's output by continually re-optimizing with a fixed-time horizon at each successive time step. 
However, as the two agents approach each other, this fixed horizon becomes restrictive. Suppose the horizon is three seconds in the future. Placing the interaction term perpetually at the fixed time horizon means that the model will always want to interact exactly three seconds in the future, independent of where it finds itself, leading to an exponential slowdown in its behavior.


We counteract the slowdown by adding an additional distance-based reward term weighted by $ \lambda_{\text{reward}}$ to the robot-agent collaboration cost defined in Eq. \ref{eq:robot-agent-collaboration-cost} at every point on the trajectory.
\begin{eqnarray}
c(\vect{\xi}^\text{R}, \vect{\xi}^\text{A}) = c(\vect{t}_T^\text{R}, \vect{t}_T^\text{A}) + \lambda_{\text{reward}} \sum_{i=0}^T r(\vect{t}_i^\text{R}, \vect{t}_i^\text{A}),
\end{eqnarray}
where the reward $r(\vect{t}_i^\text{R}, \vect{t}_i^\text{A})$ at step $i$ is defined as
\begin{eqnarray} \label{eq:reward-term}
r(\vect{t}^\text{R}_i, \vect{t}^\text{A}_i) = 1 - e^{\frac{-\left\lVert \vect{t}^\text{R}_i - \vect{t}^\text{A}_i \right\rVert^2}{2\sigma^2}}.
\end{eqnarray}
This reward term can also apply to a single agent moving toward a fixed target, where 
$\norm{\vect{t}_i^{\text{Agent}} - \text{p}^{\text{Target}}}$ 
would replace $\norm{\vect{t}_i^{\text{R}} - \vect{t}_i^{\text{A}}}$.

The reward is motivated by the types of sparse rewards used in reinforcement learning \cite{pmlr-v80-riedmiller18a}. We are rewarding the agents for converging, and, since our goal is to minimize cost, we phrase reward as negative cost. Such a reward can be modeled as an upside down radial basis function over the distance between the robot's end-effector and the interacting agent, \ie, one when the two are far apart and decreasing to zero as they draw closer.

\begin{figure}[t]
\centering
\begin{subfigure}[t]{0.4\columnwidth}
    \centering
    \includegraphics[width=.9\columnwidth]{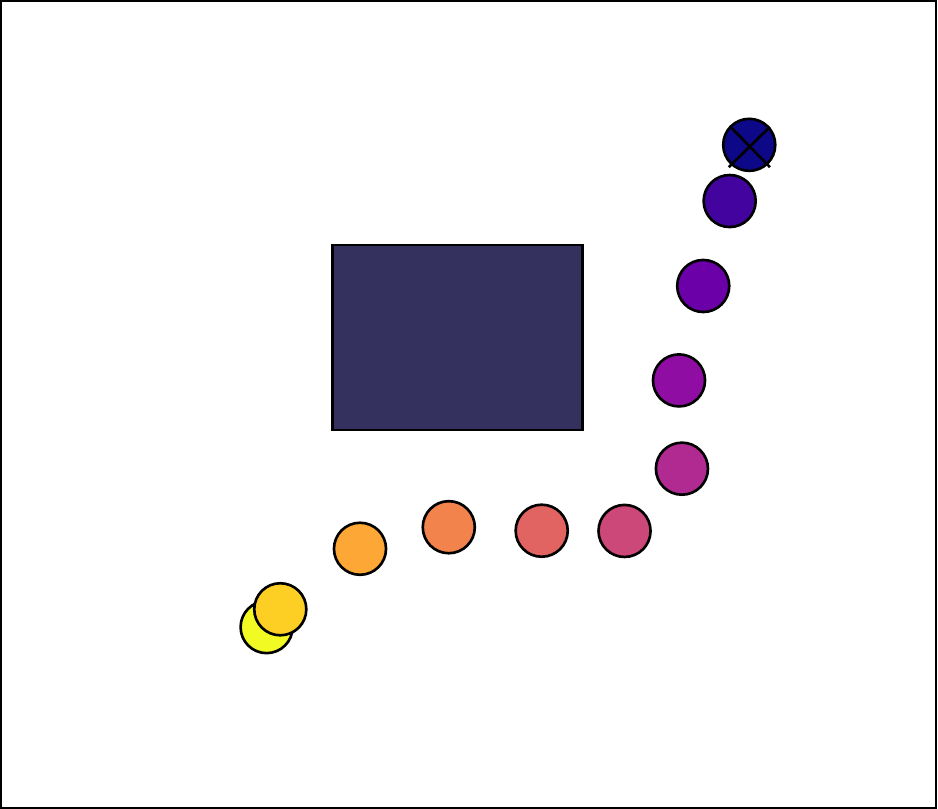}
    \captionsetup{size=footnotesize}
    \caption{Without reward term}
\end{subfigure}%
\begin{subfigure}[t]{0.4\columnwidth}
    \centering
    \includegraphics[width=.9\columnwidth]{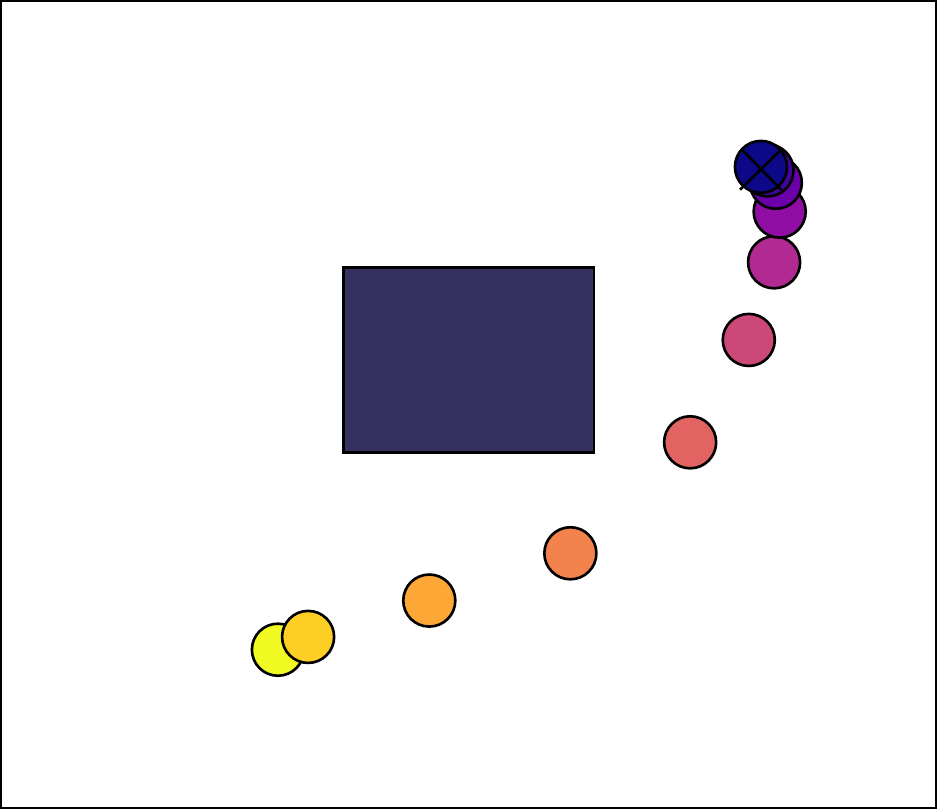}
    \captionsetup{size=footnotesize}
    \caption{With reward term}
\end{subfigure}%
\begin{subfigure}[t]{0.15\columnwidth}
    \centering
    \raisebox{-6ex}{
        \includegraphics[width=.9\columnwidth]{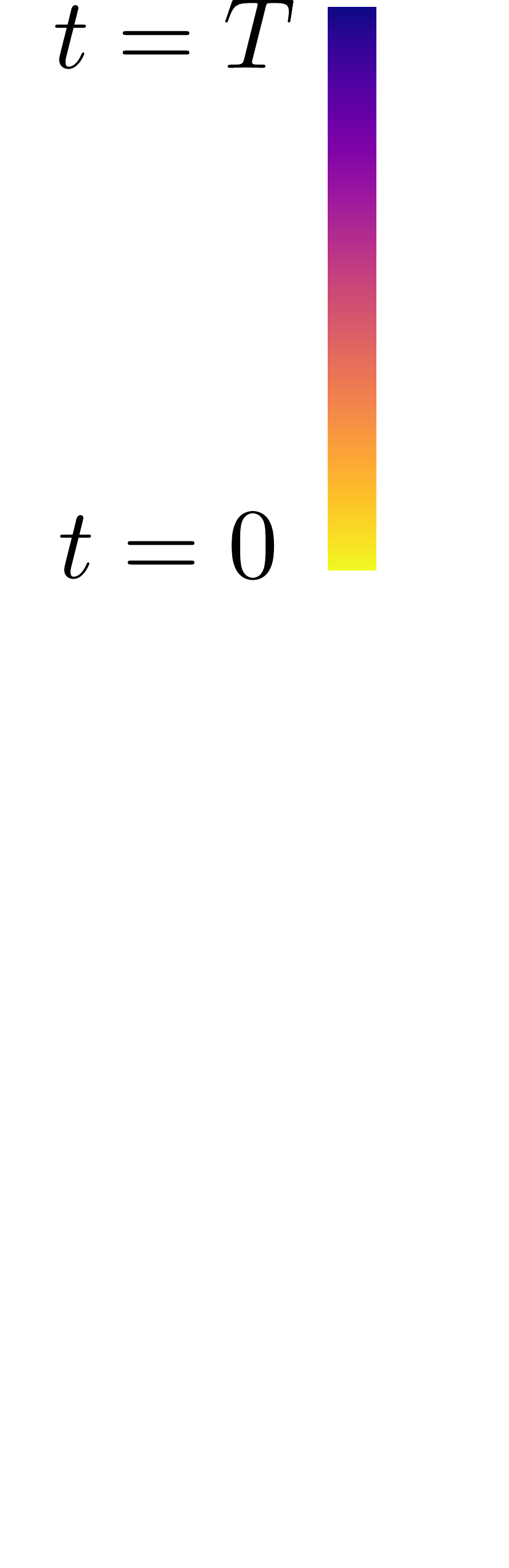}
    }
\end{subfigure}
\vspace{-4pt}
\captionsetup{size=footnotesize}
\caption{Comparison motion generation for a reach-to-point task around an obstacle both with and without the proposed sparse reward term. The dots represent subsequent positions. With each MPC step, the agent starts closer to its goal. Our reward terms encourages the agent to speed up as a function of their relative distance to the goal arrive at the goal in less time than the planning horizon.}
\vspace{-12pt}
\label{fig:sparse-reward}
\end{figure}

This rewards the robot for getting within touching distance of the interacting agent (and visa-versa), but does not penalize the pair for having to be far from each other earlier in the trajectory due to competing smoothness criteria. Since we have a fixed finite-horizon, without loss of generalization, we can shift each of these reward terms up by a constant so its minimum value is zero as given in the equation. The effect can be seen in Fig.~\ref{fig:sparse-reward}, which shows how this results in a more temporally-consistent trajectory. Specifically, in the absence of perturbations, the trajectory traced out by MPC's replan-execute loop is more consistent with the original trajectory initially planned at the first time step. 

For safety reasons, the uncontrolled agent may stop before reaching the robot. The reward term as formulated in Eq. \ref{eq:reward-term} would cause the robot to slow-down exponentially because the human policy predicts that the human will keep moving. To prevent this, we would ideally have a phase estimator that can determine when the human has stopped and switch to rewarding the robot for reaching the human's current position. As an approximation in our implementation, the human and robot are both rewarded for reaching each others' starting points, as determined at the beginning of each MPC step. 

With this shift upward, our formulation of the reward (as cost) becomes identical to the Welsch robust estimator \cite{Holland1977RobustRU}. Although the reward is not a nonlinear least squares term, it can be minimized using a form of iteratively re-weighted least-squares \cite{Zhang1997ParameterET} using weights given by the Radial Basis Function (RBF) $w(r) = e^{\frac{-r^2}{2\sigma^2}}$
where $r = \|\vect{t}^\text{R}_i - \vect{t}^\text{A}_i\|$ in this case. An implementation would replace these Welsch robust estimator objective terms with weighted least squares terms of the form $w r^2 = w\|\vect{t}^\text{R}_i - \vect{t}^\text{A}_i\|^2$, and re-evaluating the weight $w$ after each subproblem has converged. 
 
These reward terms reward the system for reaching the interaction point early. As above, associated with reaching the interaction point should be a velocity penalty bringing the system to a stop.
Whereas before, it sufficed to add just a single velocity penalty to the terminal potential (stop at the end) we now must also add intermediate velocity penalties preparing for the possibility of stopping early.

\section{Implementation Details} \label{sec:implementation}
Similar to~\cite{Mukadam2018ContinuoustimeGP}, we used GTSAM as a fast optimizer to minimize the cost at each MPC-step. We used the real-time body tracking SDK on the Microsoft Azure Kinect DK to obtain human pose estimates, and we are able to run our algorithm to perform a human handover in real time on a Franka Emika Panda arm. 


On a workstation with an \textit{3.4ghz} Intel\textsuperscript{\textcopyright} Core\textsuperscript{TM} \textit{i7}  and \textit{32GB} of RAM running Ubuntu 18.04, we obtained poses at a rate of $30$hz. We use DART \cite{Schmidt2014DARTDA} to calibrate the robot configuration into the Azure frame, so we can obtain both human and robot starting positions at each MPC step. 


We run both our optimizer and DART on the same workstation running Ubuntu 16.04 and equipped with an \textit{3.7ghz} Intel\textsuperscript{\textcopyright} Core\textsuperscript{TM} and \textit{32GB} of RAM. We obtain DART's positional estimates at $10hz$, and we are able to run our optimizer with Levenberg-Marquardt between $7$hz and $8$hz.

When the Franka is within a minimum threshold--we used $10$cm--it engages the gripper and tries to grasp the object. If it misses and closes all the way, the gripper re-opens and the planner resumes trying to engage in the handover until it succeeds. We found the robot to miss the handover when the human moves too quickly for the body tracker to maintain a stable estimate. 
 \label{sec:method}


\section{Experiments} \label{sec:experiments}
We ran a set of experiments exploring: (1) How do our method's generated trajectories compare to those produced by baseline methods? (2) How robust is our algorithm to noisy sensors? and (3) Can our proposed method be used in a real world setting?

\begin{table}[!htb]
\captionsetup{size=footnotesize}
\caption{Algorithmic benchmarks ($\uparrow$ denotes higher is better and $\downarrow$ denotes lower is better): our algorithm is best able to approximate the timing of the uncontrolled agent. 
The attractor-based algorithm produces trajectories with significantly greater acceleration and jerk than both the robot-only and our algorithm. 
The robot-only algorithm outperforms ours by a small margin in reducing acceleration and jerk, but at the cost of producing much longer trajectories.}
\setlength\tabcolsep{0pt} 
\footnotesize\centering
\begin{tabular*}{\columnwidth}{@{\extracolsep{\fill}}lrcccr}
\toprule
  Metric & & Robot only & Attractor & Ours \\
\midrule
  Handover Time (Normalized) & $\downarrow$ & 1.33 $\pm$ 0.27 & 1.30 $\pm$ 0.26 & \textbf{1.20 $\pm$ 0.26} \\
  Trajectory Length Error & $\downarrow$    & 0.35 $\pm$ 0.27 & 0.37 $\pm$  0.29 & \textbf{0.27 $\pm$ 0.22} \\
  Acceleration ($cm/s^2$) & $\downarrow$      & \textbf{4.33 $\pm$ 1.96} & 7.83 $\pm$ 3.96 & 4.72 $\pm$ 1.37 \\
  Jerk $\mu$ ($cm/s^3$)  & $\downarrow$       & \textbf{6.25 $\pm$ 2.99} & 10.06 $\pm$ 5.55 & 6.36 $\pm$ 1.88  \\
\bottomrule
\end{tabular*}
\label{Tab:evaluation}
\end{table}

Our algorithm predicts the motion and dynamics of the uncontrolled agent and reacts accordingly. In order to evaluate each component, we benchmarked our algorithm against two different baselines:

\textbf{Robot only}: A planner that only accounts for the Euclidean position of the uncontrolled agent. At each time step, the robot optimizes a trajectory around any obstacles to match its end effector position with the other agent's end effector position. 
    
\textbf{Attractor}: A planner that applies an attractor-based policy to both end effectors. This policy assumes the two arms will move toward each other at each time step. When obstacles are present, they act as repellent forces, opposing the attracting force. We implemented this method by using our same algorithm with a very short time horizon, \ie, $T=5$, which is the shortest trajectory supported by our low-level controller.

See our video submission for an example of the simulated environment. The robot and uncontrolled agent start on opposite sides of a non-convex obstacle, the position and shape of which we randomized for each trial. To evaluate the algorithms without bias, we independently planned the uncontrolled trajectory to go from a randomized location on the opposing side of the obstacle to a randomized point in the robot's reachable space, while also avoiding the obstacles. We accomplish this by minimizing our same velocity, obstacle-avoidance, and acceleration costs for the hand, while also adding a cost term with high $\lambda$ to constrain the hand to our randomly chosen start and end positions, as in \cite{Mukadam2018ContinuoustimeGP}. We augmented the plans with noise drawn from a uniform distribution to de-bias the uncontrolled motion from the planner.

We then replayed the uncontrolled trajectory for each robot policy. At the end of the uncontrolled trajectory, the agent pauses and waits for the robot. We modeled the agent as a ball with a $10$cm radius and as such, we considered a trial to be successful if the robot was able to plan a trajectory where its end effector was within $10$cm of the agent in under twice the uncontrolled trajectory length.

We ran 300 trials. Qualitatively, we observed that the randomized obstacle and randomized uncontrolled trajectory led to many ill-formed handovers, such as ones where the uncontrolled trajectory goes through the obstacles. 
The robot-only algorithm best handled these ill-formed trajectories because it is able to ignore whether the human is in an incorrect configuration. 
It succeeded in $62\%$ of the trials. 
Our algorithm succeeded in $57\%$ of the trials and the attractor policy succeeded in $43\%$ of the trials.
It is important to note that our framework assumes that uncontrolled agents are co-operative and reactive. 
To fairly compare the three policies, we used identical trajectories for the uncontrolled agent, but this precluded the possibility of the uncontrolled agent's policy being reactive to the robot and therefore violates our cooperative assumption. With a cooperative partner, we expect our policy to outperform the robot-only policy in success rate.

We evaluated the algorithms on the set of trials on which they mutually succeeded and adopted four different metrics, \ie, \textit{Success rate, trajectory length error, acceleration}, and \textit{jerk}, for evaluation. We define the \textit{trajectory length error} as $\left|1 - T_{\text{success}}/T_\text{uncontrolled}\right|$ where $T_\text{success}$ is the time it takes to finish a successful action and $T_\text{uncontrolled}$ is the length of the uncontrolled trajectory.
Quantitative results are in Table~\ref{Tab:evaluation}.

Both qualitatively and quantitatively, we saw that the \textit{attractor} algorithm leads the robot to jerk heavily when the agent's path around the obstacle is non-obvious. Meanwhile, the \textit{robot-only} planner tends to wait to move until the path around the obstacle is unobstructed, leading it to take longer to reach the agent. Our algorithm is able to smoothly predict the agents path. We also saw our algorithm produces lower \textit{trajectory length error} than the others--meaning our algorithm is better able to match the length of the uncontrolled trajectory. See our video for a demonstration.



\begin{table}[bt]
\def\tabcolsep{1.3em}
\def\arraystretch{1.5}

\captionsetup{size=footnotesize}
\caption{Robustness metrics: we evaluate our robustness to measurement noise by determining, for a given amount of measurement noise, the percentage of handovers that can be completed within twice the time of the uncontrolled trajectory}

\centering
\begin{tabular}{l ccccc}
\toprule
Noise $\sigma$ (cm) & 2   & 5   & 7  & 10 & 15 \\ \midrule
\% Successful       & 100 & 100 & 98 & 86 & 66 \\ 
\bottomrule
\end{tabular}
\label{Tab:robustness}
\vspace{-10pt}
\end{table}

We also evaluated our algorithm's robustness to measurement noise. In our real-robot experiments, we observed that our planner failed when the calibration and/or body tracker were misaligned. To measure this, we planned a randomized uncontrolled trajectory in the same fashion as before. However, we also introduced Gaussian noise with increasing $\sigma$ into the robot's perception of the trajectory. We ran $50$ trials and measured how often the robot could intersect the agent at its \textit{actual} location within twice the uncontrolled trajectory length. Results with varying $\sigma$ are in Table \ref{Tab:robustness}.

As shown in Figure \ref{fig:cover}, we are able to run our algorithm on a real robot using the setup described in Section \ref{sec:implementation}. See our video for more examples.

\section{Conclusion and Future Work} \label{sec:conclusions}

We proposed an MPC approach for multi-agent collaboration problems that simultaneously optimizes motion plans for a robot and an (uncontrolled) human in order to enable coordination on cooperative tasks, with an application to human-robot handovers in obstacle-rich environments. We presented a novel theoretical framework and demonstrated its effectiveness through both simulated and real-robot experiments. This framework assumes access to a model for the human collaborator, and future work might learn such a model from data, for example, via Inverse Optimal Control~\cite{ZiebartHumanMaxEntIOC2009}, which has been successfully applied to motion prediction in the past~\cite{kitani2012activity}. In addition, our approach cannot generate longer-term plans across manipulations, as in~\cite{toussaint2017multi}; in the future, we will develop approaches which maintain reactivity but allow for switching between discrete modes, such as when the human is waiting for the robot at the end of a handover, possibly via the Robust Logical-Dynamical Systems formalism~\cite{paxton2019representing}.
We also intend to apply our formal system to other coordination problems explored in the literature, such as motion in a crowd~\cite{bai2015intention} and camera control~\cite{rakita2018autonomous, Bonatti2019TowardsAR}.


\bibliographystyle{IEEEtran}
\bibliography{references}
\end{document}